\documentclass[runningheads]{llncs}
\usepackage[T1]{fontenc}
\usepackage{graphicx}
\usepackage{graphicx}

\usepackage{tikz}
\usepackage{comment}
\usepackage{amsmath,amssymb}
\usepackage{color}
\usepackage{algorithm}
\usepackage{algpseudocode}
\usepackage{url}
\usepackage{booktabs}
\usepackage{multirow}
\usepackage{array}
\usepackage[accsupp]{axessibility}
\usepackage[utf8]{inputenc}
\usepackage[english]{babel}
\usepackage{float}

\usepackage{hyperref}
\usepackage{amsmath}
\begin{document}
\footnotesize
\title{Pandar128 dataset for lane line detection\thanks{The work has been supported by the grant of the University of West Bohemia, project No. SGS-2025-011. Computational resources were provided by the e-INFRA CZ project (ID:90254), supported by the Ministry of Education, Youth and Sports of the Czech Republic. \url{https://github.com/filipberanek/lane_lines_dataset}}}

%
%
%
\author{
Filip Beránek\inst{1}\orcidID{0009-0002-3563-3097} \and
Václav Diviš\inst{1}\orcidID{0000-0001-9935-7824} \and 
Ivan Gruber \inst{1}\orcidID{0000-0003-2333-433X}
}

\authorrunning{Filip Beránek et al.}

\institute{Department of Cybernetics and New Technologies for the Information Society, Technická 8, 301 00 Plzeň, Czech Republic 
\email{bionothi@ntis.zcu.cz}\\}
%
\maketitle              
\begin{abstract}
We present Pandar128, the largest public dataset for lane line detection using a 128-beam LiDAR. It contains over 52,000 camera frames and 34,000 LiDAR scans, captured in diverse real-world conditions in Germany. The dataset includes full sensor calibration (intrinsics, extrinsics) and synchronized odometry, supporting tasks such as projection, fusion, and temporal modeling.

To complement the dataset, we also introduce SimpleLidarLane, a light\-weight baseline method for lane line reconstruction that combines BEV segmentation, clustering, and polyline fitting. Despite its simplicity, our method achieves strong performance under challenging various conditions (e.g., rain, sparse returns), showing that modular pipelines paired with high-quality data and principled evaluation can compete with more complex approaches.

Furthermore, to address the lack of standardized evaluation, we propose a novel polyline-based metric — Interpolation-Aware Matching F1 (IAM-F1) — that employs interpolation-aware lateral matching in BEV space.

All data and code are publicly released to support reproducibility in LiDAR-based lane detection.

\end{abstract}
%
%
%

%
%
%
\section{Introduction}

In pursuit of safer and more comfortable roadways, automotive manufacturers are developing the Advanced Driver Assistance Systems (ADAS)~\cite{ADAS_Review}. These systems, ranging from Adaptive Cruise Control (ACC) to Lane-Keeping Assist (LKA), represent a critical step towards fully autonomous vehicles~\cite{Autonomous_cars}.

LKA is one of the critical systems for autonomous vehicles which is required by law in the European Union (EU) to be present in all new cars sold in EU~\cite{LaneReqEU} starting from July 2024. The task of LKA is to detect accurate position and curvature of the ego lane and adjacent lanes, which is provided as an input for path planning of the car. Given its relevance to functional safety, the system must operate reliably across all conditions, including day and night, heavy rain, fog, and dense traffic.

The early methods for lane detection relied on classical image processing techniques applied to camera data, such as edge detection followed by line fitting~\cite{DoubleLaneDetectionHough,RobustLaneDetection,DeepLaneDetectionStructure}.
Later on, approaches based on deep neural networks were developed, which improved the accuracy and robustness~\cite{condlanenettoptodownlanedetection,spatialdeepspatialcnn,eyeslanerealtimeattentionguided}. Unfortunately, the number of algorithms is significantly exceeding the amount of evaluating datasets. Up until now, there were only 3 datasets for benchmarking of lane detection. Camera-based datasets CULane~\cite{spatialdeepspatialcnn} and TuSimple~\cite{TuSimple} and Light Detection and Ranging (LiDAR) based dataset K-lane~\cite{klane}.

Despite these advances, camera-based approaches still face several limitations that hinder their robustness in real-world applications. First, their performance is highly sensitive to lighting conditions—low light, glare, strong shadows, or nighttime scenarios can significantly reduce detection accuracy~\cite{klane,spatialdeepspatialcnn}. Second, projecting detected lanes from the camera image plane into the vehicle's 3D coordinate system can introduce geometric distortions, particularly on curved or sloped roads~\cite{bai2019deepmultisensorlanedetection}. These complications can be addressed by using LiDAR sensors, which inherently capture three-dimensional spatial information and exhibit robustness to challenging lighting conditions, including low light and glare~\cite{paek2022rowwiselidarlanedetection,klane,liDARSLAMrobust2024}.

To support the community in developing and testing LKA applications, we publish a new LiDAR-based dataset and hence contribute to the overall LKA development process as follows:
\begin{itemize}
\item We present Pandar128, the first lane detection dataset collected with a 128-beam LiDAR.
\item Pandar128 is the largest dataset for lane line detection, containing 52200 images and 34829 LiDAR frames. 
\item For the first time, we include vehicle odometry together with full camera and LiDAR calibration parameters (intrinsics, extrinsics, and distortion), enabling accurate projection and sensor fusion.
\item Furthermore, we present a baseline method for LiDAR detection combining deep learning and clustering.
\item We introduce a novel evaluation metric for polyline-based lane detection that computes average point-to-line distance with tolerance-aware matching.
\end{itemize}
\section{Related Work}

\subsection{Datasets}
A comprehensive overview of ADAS-related datasets is provided in~\cite{sarker2024comprehensive}. However, only a small subset of those datasets include LiDAR point clouds, and even fewer provide annotations for lane markings—an essential component for autonomous driving research.
To the best of our knowledge, only three publicly available datasets offer labeled lane markings in LiDAR point clouds: Kaist-Lane (K-Lane)~\cite{klane}, LiSV-3DLane~\cite{zhao2024advancements}, and Waymo Open Dataset (Waymo)~\cite{sun2020scalability}, the latter updated with semantic annotations in March 2024. 
\begin{itemize}
    \item \textbf{Waymo}~\cite{sun2020scalability} provides 3D semantic-segmentation labels for each LiDAR frame. The dataset includes a "Lane Marker" class among 23 semantic labels, covering painted road stripes such as solid and dashed dividing lines. Waymo collected 230k LiDAR frames labeled for semantic classes, under different weather conditions.
    \item \textbf{K-Lane}, introduced by Bai et al.~\cite{klane}, is a LiDAR-based end-to-end framework that predicts lane boundaries directly in the bird’s-eye view (BEV) space. Alongside the method, the authors also released a dataset containing over 15,000 annotated frames with up to six labeled lanes per scene. The data covers a wide range of road and traffic conditions, including varying occlusion levels, both daytime and nighttime driving, curved road segments, and complex topologies such as merging and diverging lanes.
    \item \textbf{LiSV-3DLane}, developed by Zhao et al.~\cite{Zhao2023Advancements3DLane}, is a large-scale 3D lane dataset comprising 20,000 surround-view LiDAR frames with detailed semantic annotations. Unlike prior datasets restricted to a frontal field of view, LiSV-3DLane offers full 360-degree coverage around the ego vehicle, allowing the capture of complex lane geometries in both urban and highway environments.
\end{itemize}

\subsection{Lane Detection}

In recent years, LiDAR-based lane detection has emerged as a robust alternative to camera-based methods, particularly under adverse weather and lighting conditions ~\cite{paek2022rowwiselidarlanedetection,zywanowski2020lidarloop,zhang2023sensorsurvey}. Several approaches have been proposed to extract lane markings directly from 3D point clouds using deep learning or hybrid techniques. These methods differ significantly in terms of input representation (raw point cloud vs. bird’s-eye view projection), network architecture, and output format (segmentation maps, polylines, or continuous surfaces).

Bai et al.~\cite{klane} proposed an end-to-end LiDAR-based system called K-Lane, which directly predicts lane boundaries in bird’s-eye view using a deep neural network. Their method reports an F1-score of 82.1\% on the K-Lane dataset, combining lane localization accuracy with classification quality and serves as an important benchmark in the field. The authors also introduce an instance-aware detection strategy, allowing their model to detect multiple lanes in complex urban scenarios.

Zhao et al.~\cite{Zhao2023Advancements3DLane} introduced LiLaDet, a dual-path network that jointly performs BEV segmentation and 3D spatial regression. This architecture enables accurate localization of lane lines in world coordinates and outperforms previous state-of-the-art methods. Their approach was extensively evaluated on the LiSV-3DLane dataset, which provides a challenging and diverse benchmark for full-surround LiDAR-based lane detection.

Other recent works, such as~\cite{Zhao2023Advancements3DLane,certad2025roadmarkingssegmentationlidar}, further explore the use of ray-based matching and multi-task learning to improve detection robustness and label efficiency. A broader review of LiDAR-based road structure extraction techniques is provided in~\cite{lidarbenchmarkMLS}, which positions lane detection as a critical application within the autonomous driving stack.

\section{Data}
    While the majority of existing lane detection datasets rely on camera images, our work introduces the first large-scale dataset based on high-resolution LiDAR. To the best of our knowledge, it is not only the largest publicly available dataset of its kind but also the first to include comprehensive vehicle odometry, sensor intrinsics, and extrinsics parameters. These additional calibration parameters and odometry data enable a wide range of advanced research directions beyond single-frame perception, such as multi-sensor fusion, temporal accumulation of point clouds, or map-level lane modeling. In this section, we describe the data collection setup, annotation process, and provide a detailed overview of the dataset’s structure and statistics.
    
    \subsection{Sensor Setup}
        Our sensor configuration was inspired by the Woodscape dataset~\cite{Woodscape}, which served as a reference platform for multi-modal data acquisition. 
        
        
        A high-resolution camera (2896×1876 pixels) is mounted behind the windshield, operating at 30 frames per second (FPS) or 15 FPS depending on environment. In rainy conditions, lower frame rate (15 FPS) was used to ensure higher image quality and reduce motion blur. The camera provides a 120° horizontal and 73° vertical field of view. All intrinsic and extrinsic calibration parameters, including distortion coefficients, are included in the dataset.
        
        The LiDAR sensor is a Pandar128 LiDAR unit, mounted on the roof. This LiDAR provides full 360° coverage with a vertical FOV of 40°, operating at 20 FPS. The sensor has a ranging capability of up to 200 meters at 10\% reflectivity. 
        Compared to other widely used datasets such as KITTI-360~\cite{kitti360} and K-Lane~\cite{klane}, which rely on 64-beam LiDARs, the use of 128 beams allows for denser and more detailed scene reconstruction.
        

        The final component of the sensor suite is a high-precision GPS/IMU system providing 20 Hz odometry. Instead of absolute coordinates, we store relative transformation matrices between consecutive frames, enabling motion compensation, temporal fusion, and map-level reconstruction. Although raw GPS data is omitted, the odometry ensures sufficient spatial consistency for multi-frame alignment and time-aware modeling.

        \begin{figure}[h]
            \centering
            \includegraphics[width=\textwidth]{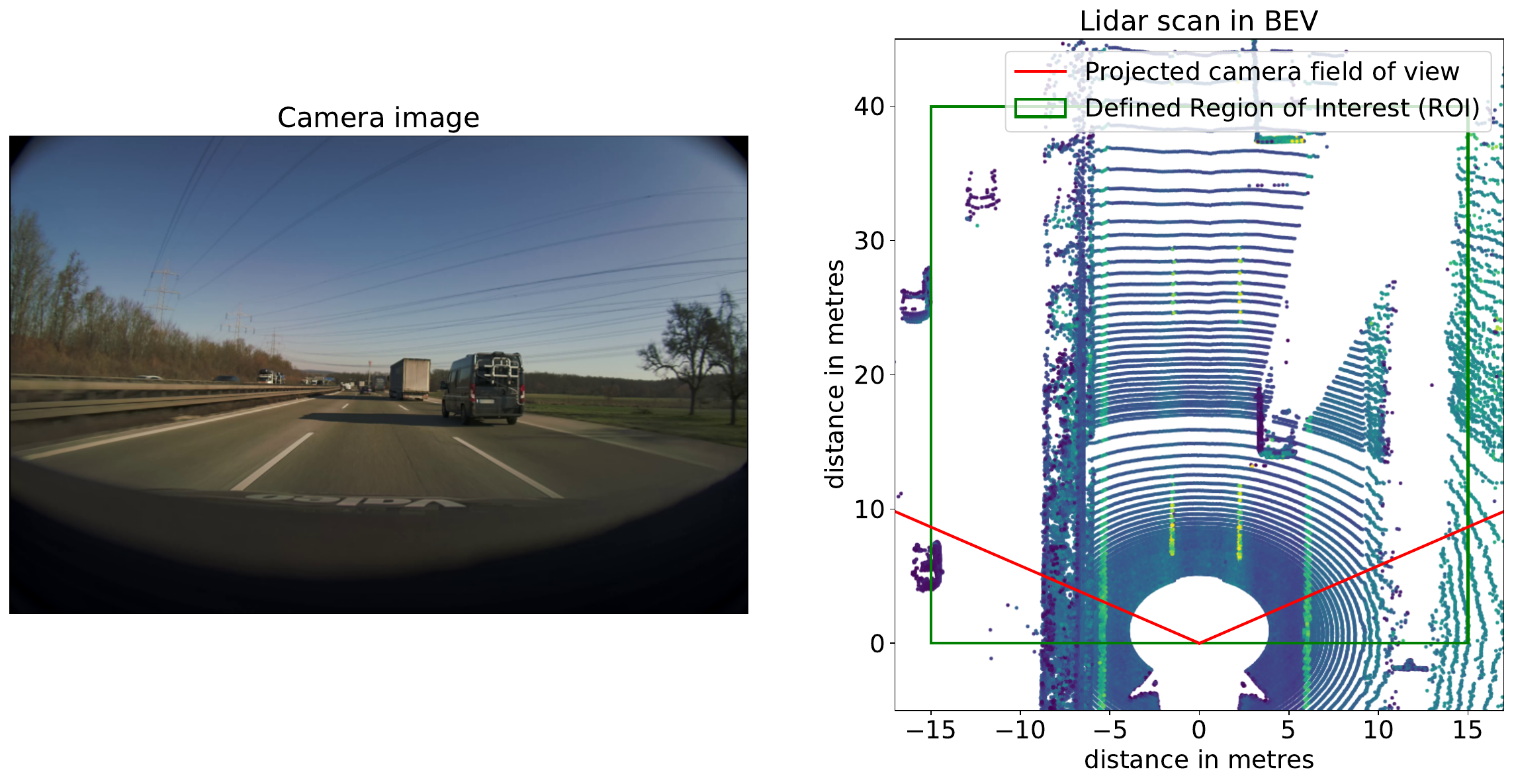}
            \caption{Image from the front-facing camera (left) and the corresponding LiDAR scan (right).}
            \label{fig:frame_raw_camera_and_scan}
        \end{figure}
        
        Fig.~\ref{fig:frame_raw_camera_and_scan} illustrates a synchronized snapshot from our dataset, combining the front-facing camera view with the corresponding LiDAR scan in a BEV. The camera’s horizontal field of view is overlaid on the BEV point cloud, highlighting the spatial alignment between modalities. For all downstream tasks, we define a consistent region of interest (ROI) matching K-Lane~\cite{klane}: from 0 to 40 meters ahead of the vehicle and ±15 meters laterally.

    \subsection{Dataset collection}
    
        Our dataset comprises of 29 driving traces (sequences), each representing a 60-second segment of continuous driving. Each trace includes approximately 1,800 video frames and cca 1,200 LiDAR scans with synchronized odometry data. In total, the dataset provides 46,802 image frames and 34,831 LiDAR and odometry frames.

        A comparison with other publicly available datasets is provided in Table~\ref{tab:datasets_comparison}.

        \begin{table}[!h]
            \centering
            \small
            \begin{tabular}{l|c|c|@{}c@{}|@{}c@{}|c}
            \toprule
            \textbf{Dataset} & \textbf{Frames} & \textbf{Beams} & \textbf{Situations} & 
            \begin{tabular}[t]{@{}c@{}}
            \textbf{Daytime} \\ \textbf{Weather} \end{tabular} & 
            \textbf{Perspective [°]} \\
            \midrule
            \cite{sun2020scalability} Waymo 
            & 230k & 64 & \begin{tabular}[t]{@{}c@{}} urban, \\ highway \end{tabular} 
            & \begin{tabular}[t]{@{}c@{}} day, night \\ rain, snow, fog \end{tabular}
            & 360 \\
            \addlinespace
            \cite{Zhao2023Advancements3DLane} LiSV-3DLane  
            & 20k & 64 & \begin{tabular}[t]{@{}c@{}} urban, \\ highway \end{tabular}
            & \begin{tabular}[t]{@{}c@{}} day, night \end{tabular}
            & 360 \\
            \addlinespace
            \cite{klane} K-Lane
            & 15k & 64 & \begin{tabular}[t]{@{}c@{}} urban, \\ highway \end{tabular} 
            & \begin{tabular}[t]{@{}c@{}} day, night \\ rain, fog \end{tabular}
            & 180 \\
            \addlinespace
            \textbf{Pandar128 (ours)} 
            & \textbf{35k} & \textbf{128} & \begin{tabular}[t]{@{}c@{}} \textbf{urban}, \\ \textbf{highway} \end{tabular} 
            & \textbf{day, rain} 
            & \textbf{360} \\
            \bottomrule
            \end{tabular}
        \caption{Comparison of publicly available lane line detection datasets.}
        \label{tab:datasets_comparison}
        \end{table}

        All data was collected in Germany, with a strong focus on highway driving under sunny and low-traffic conditions. While these scenarios dominate the dataset, it also includes a smaller number of traces featuring rain, construction zones, and mid-traffic conditions, enabling initial exploration of robustness in more challenging settings.
        
        Key characteristics of the dataset based on trace level can be seen in Table~\ref{tab:key_characteristic}.

        \begin{table}[!h]
            \centering
            \label{tab:dataset_statistics_compact}
            \begin{tabular}{c|c|c||c|c|c}
                \toprule
                \multicolumn{3}{c||}{\textbf{Road Type}} &
                \multicolumn{3}{c}{\textbf{Weather Type}} \\
                \cmidrule(r){1-3} \cmidrule(r){4-6}
                City & Expressway & Highway &
                Sunny & Cloudy & Rainy \\
                3 (10.3\%) & 2 (6.9\%) & 24 (82.8\%) &
                22 (75.9\%) & 2 (6.9\%) & 5 (17.2\%) \\
            \end{tabular}
            
            \vspace{0.3em} 
        
            \begin{tabular}{c|c||c|c}
                \midrule
                \multicolumn{2}{c||}{\textbf{Traffic Level}} &
                \multicolumn{2}{c}{\textbf{Roadwork}} \\
                \cmidrule(r){1-2} \cmidrule(r){3-4}
                Mid-traffic & Low-traffic & No const. & Const. zone \\
                7 (24.1\%) & 22 (75.9\%) & 26 (89.7\%) & 3 (10.3\%) \\
                \bottomrule
            \end{tabular}
        \caption{Distribution of scenario attributes across the dataset, broken down by road type, weather conditions, traffic density, and presence of construction zones.}
        \label{tab:key_characteristic}
        \end{table}

        

    \subsection{Annotations}
    
        We provide two complementary annotation types for lane lines:
        \begin{itemize}
            \item Semantic segmentation of the point cloud
            \item Lightweight polyline annotations
        \end{itemize}
        While semantic segmentation offers detailed point-level labels, polyline annotations defined as sparse sequences of (x, y, z) coordinates significantly reduce annotation time and cost. Furthermore, polylines are more computationally efficient for certain downstream tasks, such as lane modeling and tracking.

\begin{figure}[!h]
            \centering
            \includegraphics[width=\textwidth]{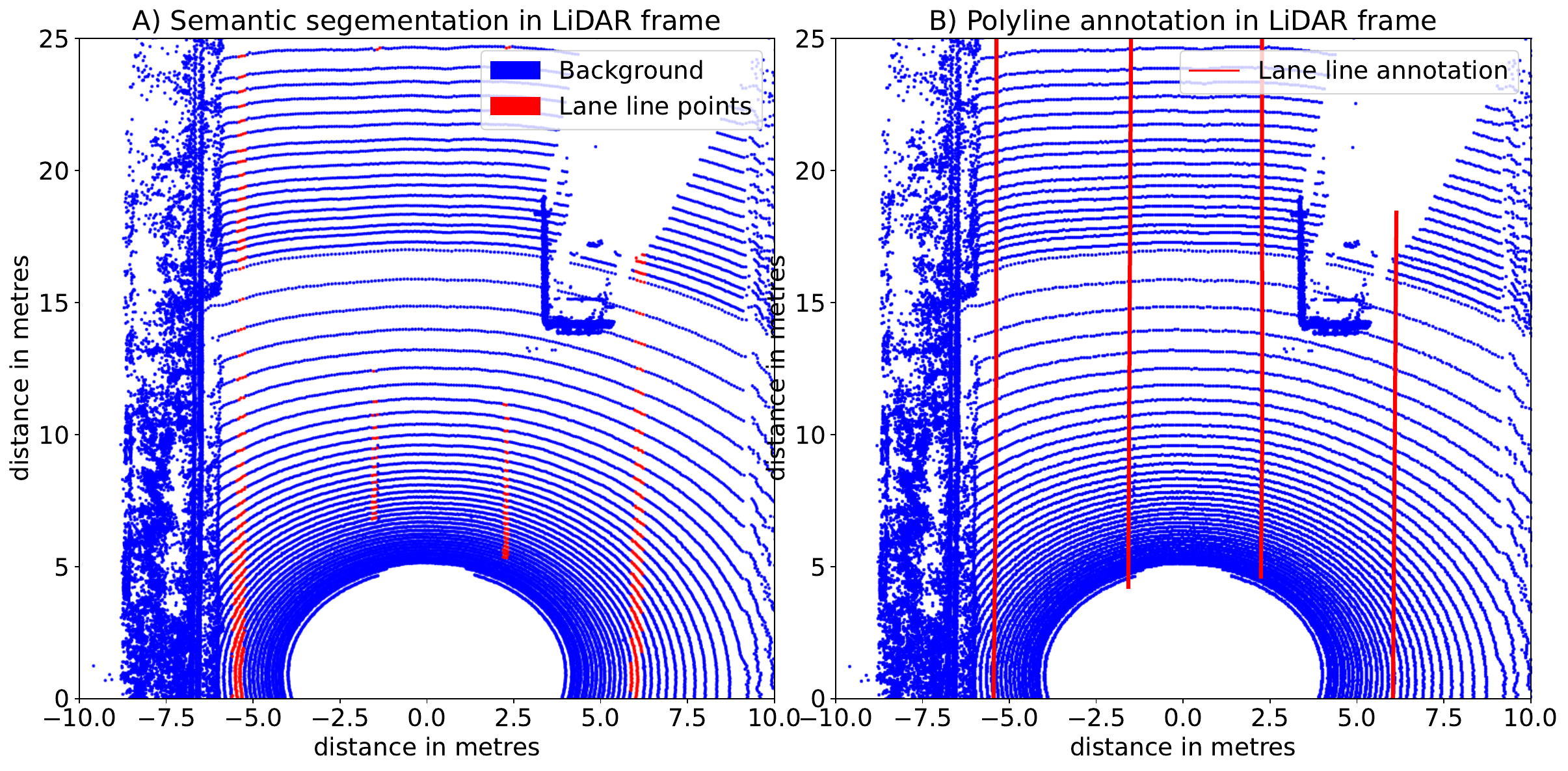}
            \caption{A) Example of point cloud semantic segmentation annotation and B) Example of polyline annotation.}
            \label{fig:annotation_example}
        \end{figure}

        The annotation guidelines were designed to support not only lane detection, but also higher-level behaviors such as vehicle cut-ins, cut-outs, and trajectory planning. 
        Semantic segmentation annotation rules are:
        \begin{itemize}
            \item Annotate all visible lane lines in the ego-lane and adjacent lanes in the ego direction.
            \item Do not annotate opposite-direction lanes when separated by a physical barrier.
            \item Include visible parts of exit and acceleration lanes.
            \item Do not annotate occluded or non-visible lane lines.
            \item Annotate white lane line with label 1, yellow lane line with label 2 and background as 0. 
        \end{itemize}
        An example of semantic segmentation annotation can be seen in Fig.~\ref{fig:annotation_example}.
        
        Polyline annotations are represented as ordered lists of (x, y, z) coordinates in 3D space, allowing fast and lightweight labeling while preserving key structural information. These polylines can optionally be rasterized into a meshgrid format following the approach in K-Lane~\cite{klane} by mapping nearby mesh cells to the polyline trajectory. An example of our a polyline annotation using (x, y, z) coordinates is shown in Fig.~\ref{fig:annotation_example}. 
        The same annotation rules apply as in the case of semantic segmentation, with the additional constraint that polylines must lie strictly within the physical extent of the lane line, ensuring geometric consistency.

\begin{figure}[!h]
        \centering
        \includegraphics[width=\textwidth]{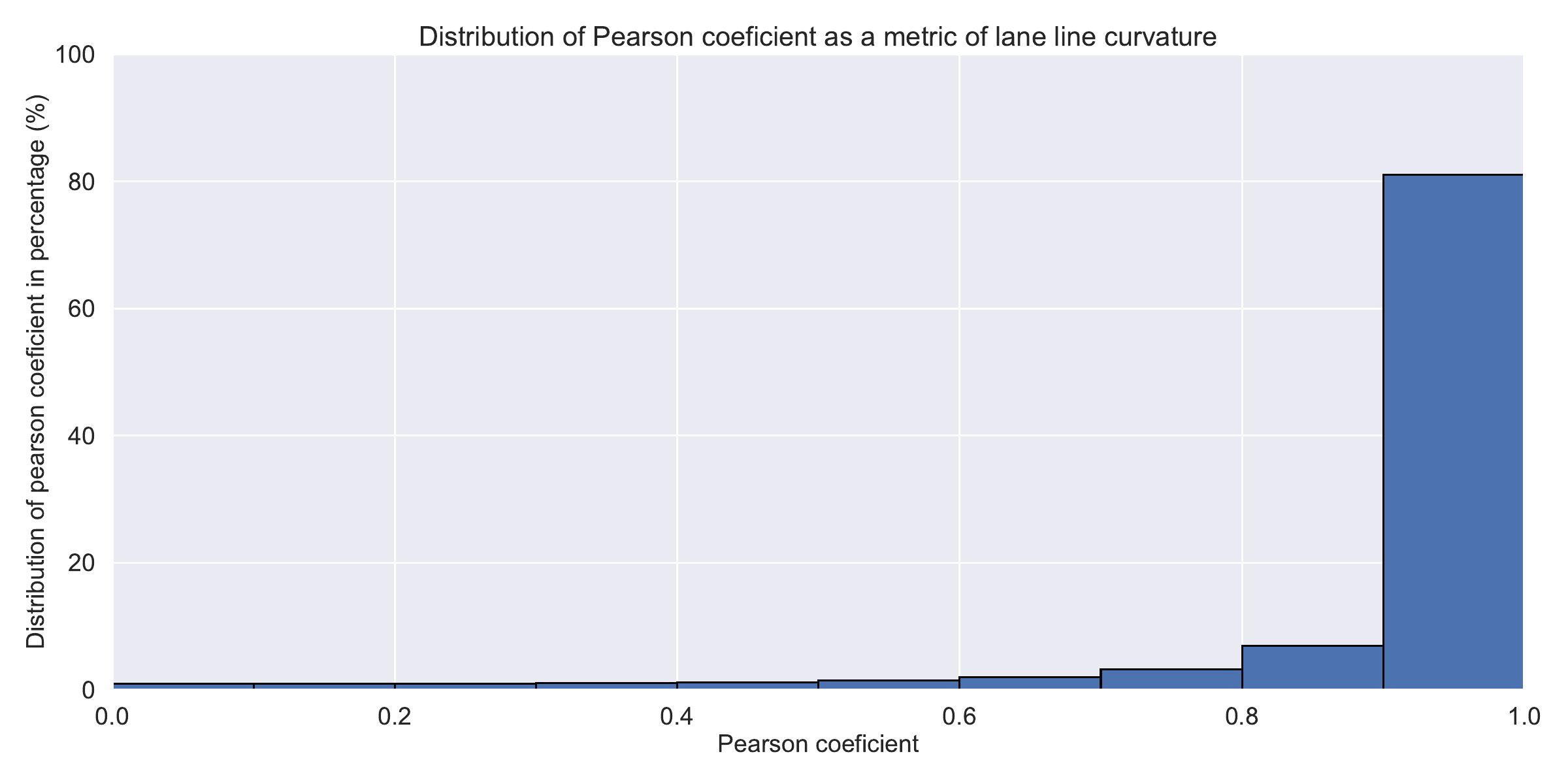}
        \caption{Distribution of curvature (measured via Pearson coefficient) of annotated lane lines across frames.}
        \label{fig:lane_line_curviture}
        \end{figure}

        We assessed lane curvature by computing the Pearson correlation between x and y coordinates of each polyline. Values near 1 indicate straight segments, while lower values reflect higher curvature or complex geometries. Aggregated per frame distribution is shown in Fig.~\ref{fig:lane_line_curviture}.

\begin{figure}[!h]
        \centering
        \includegraphics[width=\textwidth]{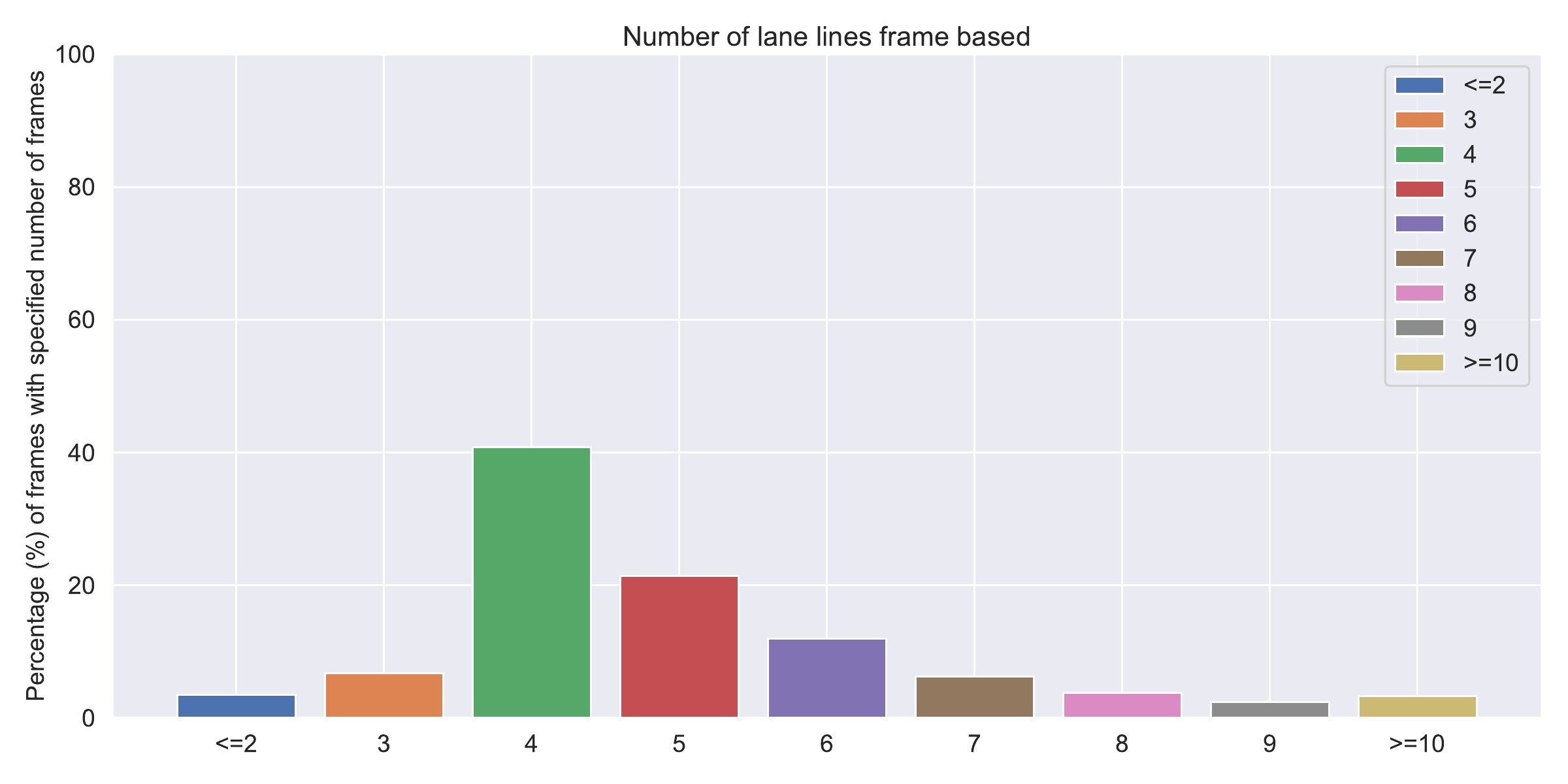}
        \caption{Distribution of the number of annotated lane lines per frame across the dataset.}
        \label{fig:number_of_lane_line_frequency}
        \end{figure}

        Scene complexity was evaluated by counting lane lines per frame and aggregating these counts across all traces. The resulting distribution (Fig.~\ref{fig:number_of_lane_line_frequency}) shows the variation in lane configurations.
        
\subsection{Data Split Strategy}
\label{subsec:data_split_strategy}

We divided our dataset, which consists of 29 driving traces, into training, validation, and test subsets with careful attention to diversity in road type, weather, traffic density, and construction zones. 

To ensure representative evaluation, both the validation and test sets contain 6 traces, each explicitly selected to include at least one challenging condition (e.g., rain, mid-traffic, construction zones, or urban driving). 

The remaining 17 traces form the training set and include at least one instance of each rare scenario, supporting generalization to diverse environments. 

This stratified split enables consistent benchmarking and promotes robust model evaluation across both common and edge-case driving conditions.

\section{Methods}
\subsection{SimpleLidarLane: Baseline Extraction Method}
\label{subsec:simple_lidar_lane}

\begin{figure}[!ht]
    \includegraphics[width=\textwidth]{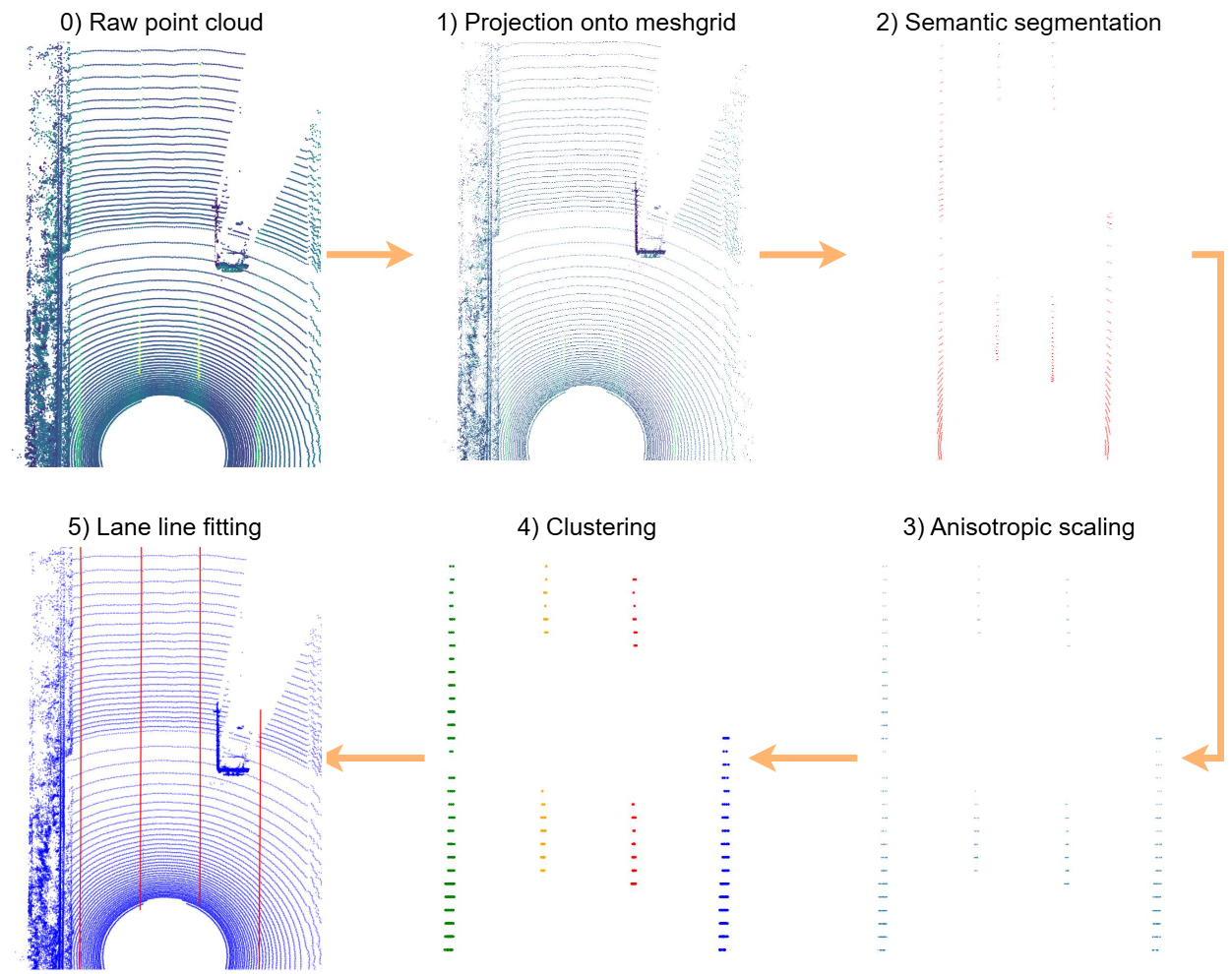}
    \caption{Architecture of the proposed SimpleLidarLane pipeline.}
    \label{fig:baseline_architecture}
\end{figure}

To provide a lightweight and interpretable baseline for lane line extraction, we propose SimpleLidarLane, a modular pipeline that combines semantic segmentation with classical post-processing techniques. Unlike end-to-end approaches such as K-Lane~\cite{klane}, our method decomposes the task into clearly defined submodules:

\begin{enumerate}
    \item Project the LiDAR point cloud onto a 2D meshgrid representing a bird's-eye view (BEV) with a field of view covering 0–40 meters in front of the vehicle and ±15 meters to the sides, which is similar to the setup used in \cite{klane}. The grid resolution is set to 5\,cm per cell in real-world units.
    \item Perform semantic segmentation using a U-Net~\cite{unet} architecture with ResNet-18 \cite{he2016deep} as backbone.
    \item Apply anisotropic scaling by elongating the forward axis and compressing the lateral axis to enhance lane separation.
    \item Clustering using DBSCAN~\cite{DBSCAN}.
    \item Lane line fitting using RANSAC~\cite{RANSAC}.
\end{enumerate}
An overview of the individual steps is illustrated in Fig.~\ref{fig:baseline_architecture}.

This decomposition enhances interpretability: if the output fails, one can directly analyze whether the issue stems from the perception stage (e.g., misclassified pixels), the clustering (e.g., overlapping lane instances), or the fitting procedure (e.g., inaccurate curves). Such isolation supports targeted debugging and facilitates systematic performance improvements. 

Compared to dense segmentation-based pipelines, our polyline representation reduces memory footprint, annotation effort, and aligns better with vector-based planning modules. While advanced methods like LiLaDet~\cite{Zhao2023Advancements3DLane,klane} achieve strong performance using deep 3D regression, and others explore multi-sensor fusion from camera and LiDAR domains~\cite{bai2019deepmultisensorlanedetection}, our design prioritizes robustness, clarity, and ease of integration. 

Ultimately, SimpleLidarLane is intended to serve as a strong and interpretable baseline against which more complex architectures can be compared. Its simplicity, modularity, and efficiency also make it suitable as a practical tool for rapid deployment and prototyping in real-world lane detection applications.

\subsection{Evaluation Metrics}

We evaluate lane line extraction using both standard metrics and a novel polyline-based metric that we introduce in this work. The full evaluation framework is composed of:

\begin{itemize}
\item \textbf{Proposed metric:} Interpolation-Aware Matching F1 (IAM-F1).
\item \textbf{Standard metrics:} semantic segmentation F1, raster-based polyline F1.
\end{itemize}

\subsubsection{Interpolation-Aware Matching F1 (IAM-F1)}

We introduce the Interpolation-Aware Matching F1 (IAM-F1), a novel metric designed to quantify the geometric alignment between predicted and ground-truth polylines in the bird’s-eye view (BEV). IAM-F1 focuses on 2D lateral accuracy, which is critical for vehicle control and path planning, and operates exclusively in the $(x, y)$ plane. The vertical ($z$) component is deliberately omitted, as lane lines are predominantly used for lateral positioning and guidance in the road plane.

Formally, let $\mathcal{G} = \{g_j = (x_j, y_j)\}_{j=1}^N$ denote a ground-truth polyline and $\mathcal{P} = \{p_k = (x_k, y_k)\}_{k=1}^M$ a predicted polyline. Both are represented as ordered sequences of 2D points in BEV space.

For each ground-truth point $g_j$, we compute the interpolated lateral position $y^\text{interp}_j$ on $\mathcal{P}$ via linear interpolation at the same longitudinal position $x_j$:

\begin{equation}
y^\text{interp}_j = \mathrm{interp}_{\mathcal{P}}(x_j), \qquad
\delta_j = |y^\text{interp}_j - y_j|
\end{equation}

If $\delta_j < \tau$ (with $\tau = 0.2\,\text{m}$), it is considered as a true positive (TP). Otherwise, it is a false negative (FN). Points outside the interpolation domain are also counted as FN.

We repeat the process in reverse: interpolating $\mathcal{G}$ at each $x_k$ from $\mathcal{P}$ to compute false positives (FP).

The metric is then computed as:

\begin{align}
\text{TP} &= \left| \left\{ g_j \mid x_j \in \text{dom}(\mathcal{P}) \wedge \delta_j < \tau \right\} \right| \\
\text{FN} &= \left| \left\{ g_j \mid x_j \notin \text{dom}(\mathcal{P}) \vee \delta_j \geq \tau \right\} \right| \\
\text{FP} &= \left| \left\{ p_k \mid x_k \notin \text{dom}(\mathcal{G}) \vee |y_k - \mathrm{interp}_{\mathcal{G}}(x_k)| \geq \tau \right\} \right|
\end{align}


In contrast to prior methods, including K-Lane~\cite{klane}, our approach adopts a stricter and explicitly defined matching threshold. IAM-F1 provides a transparent, geometry-aware metric for polyline alignment in BEV space and offers computational advantages by operating on sparse polylines instead of dense raster maps.

\begin{figure}[h]
\centering
\includegraphics[width=\textwidth]{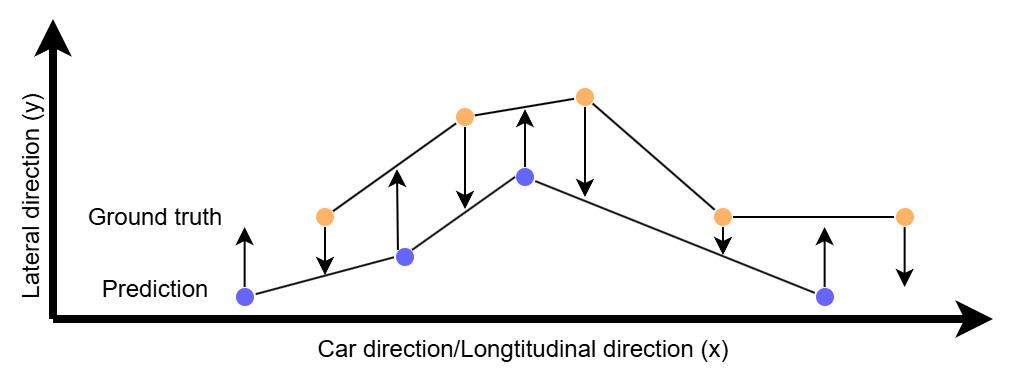}
\caption{Illustration of interpolation-aware lateral matching in IAM-F1 metric.}
\label{fig:metrics_calculation}
\end{figure}

A schematic of the interpolation and thresholding process is shown in Fig.~\ref{fig:metrics_calculation}.

This makes IAM-F1 suitable for scalable evaluation on large datasets.

\subsubsection{Standard Metrics}
In addition to our proposed IAM-F1 metric, we evaluate performance using two standard approaches for comparability with prior work:

\begin{enumerate}
    \item \textbf{Semantic Segmentation F1} — This metric computes the F1 score over meshgrid cells corresponding to annotated lane markings. It reflects the network’s ability to localize lane areas in BEV space at the segmentation level.
    
    \item \textbf{Polyline-Based Meshgrid Evaluation} — Inspired by K-Lane~\cite{klane}, we evaluate polylines by discretizing both predicted and ground-truth annotations onto a shared BEV meshgrid. Binary masks are generated to distinguish lane vs. background cells, and standard classification metrics (F1 score, precision, recall) are computed over the resulting occupancy grids. This method measures spatial agreement in a grid-based representation without requiring explicit instance matching.
    
\end{enumerate}

\section{Experiments}
\subsection{Experimental setup}
We train only the semantic segmentation component of the SimpleLidarLane pipeline, which operates on bird’s-eye view (BEV) meshgrid representations of LiDAR point clouds. The BEV grid spans 0–40 meters longitudinally and ±15 meters laterally with a resolution of 5 cm per cell similar to K-Lane\cite{klane}, as described in Section~\ref{subsec:simple_lidar_lane}.

We use a U-Net++~\cite{unetplutplus} segmentation architecture with a ResNet-18~\cite{he2016deep} encoder pretrained on ImageNet. The model outputs three semantic classes: background, white lane line, and yellow lane line.

Given the strong class imbalance (lane lines occupy only a small portion of the BEV), we employ focal loss~\cite{lin2018focallossdenseobject} with default settings to improve convergence. No data augmentations are applied.

Training is performed using the Adam optimizer ($\epsilon = 10^{-8}$) with an initial learning rate of 0.01. We apply ReduceLROnPlateau scheduling (patience = 5 epochs, $\epsilon = 0.2$) and early stopping (patience = 10 epochs) based on validation IoU. The batch size is 2 for training and 1 for validation.

The model is trained for up to 400 epochs on an NVIDIA RTX 4060 GPU with 8GB VRAM, 32 GB LPDDR5x RAM, and an AMD Ryzen 7 7840HS CPU. The split into train, validation, and test sets follows the dataset partitioning described in Section~\ref{subsec:data_split_strategy}.

\subsection{Quantitative Results}
Table~\ref{tab:f1_comparison} reports performance across six traces from our test set, evaluated using three F1-based metrics: 
\begin{itemize}
    \item Semantic segmentation F1 
    \item Rasterized polyline F1
    \item Our proposed IAM-F1.
\end{itemize}

Segmentation performs well in structured highway scenes but degrades in complex environments such as urban–highway transitions. IAM-F1 consistently yields the highest scores, confirming the benefits of explicit geometric matching.

\begin{table}[h!]
\centering
\begin{tabular}{l|c|c|c}
\toprule
 & \multicolumn{3}{c}{\textbf{F1 score}} \\
\cmidrule(lr){2-4}
\textbf{filename} & \textbf{Segmentation} & \textbf{Polyline (meshgrid)} & \textbf{Polyline (ours)} \\
\midrule
0208\_105703\_008 & 0.8434 & 0.8669 & 0.9136 \\
0208\_105703\_012 & 0.7793 & 0.8547 & 0.9093 \\
0208\_105703\_016 & 0.7419 & 0.8787 & 0.9263 \\
0307\_160641\_111 & 0.7931 & 0.8790 & 0.9093 \\
1019\_114352\_021 & 0.7139 & 0.7657 & 0.8384 \\
1110\_131951\_005 & 0.4195 & 0.6118 & 0.5945 \\
\midrule
\textbf{Average} & \textbf{0.7152} & \textbf{0.8095} & \textbf{0.8486} \\
\bottomrule
\end{tabular}
\caption{Comparison of F1 scores for segmentation, meshgrid polyline rasterization, and our final polyline metric (rounded to 4 decimal places per trace).}
\label{tab:f1_comparison}
\end{table}

Unlike raster-based metrics tied to fixed meshgrid resolution (5×5\,cm), IAM-F1 compares sparse polylines directly, avoiding quantization artifacts and enabling sharper, more efficient evaluation.




\subsection{Qualitative Evaluation}
Our SimpleLidarLane achieves an average IAM-F1 score of \textbf{80.95\%} on the test set, demonstrating consistent performance across diverse driving conditions. The method remains effective even when lane markings are barely visible due to glare, rain, or low illumination. Despite LiDAR sparsity at longer ranges, it reliably reconstructs lane lines up to 40\,m, and under adverse weather up to 25\,m.

\begin{figure}[!ht]
    \includegraphics[width=\textwidth]{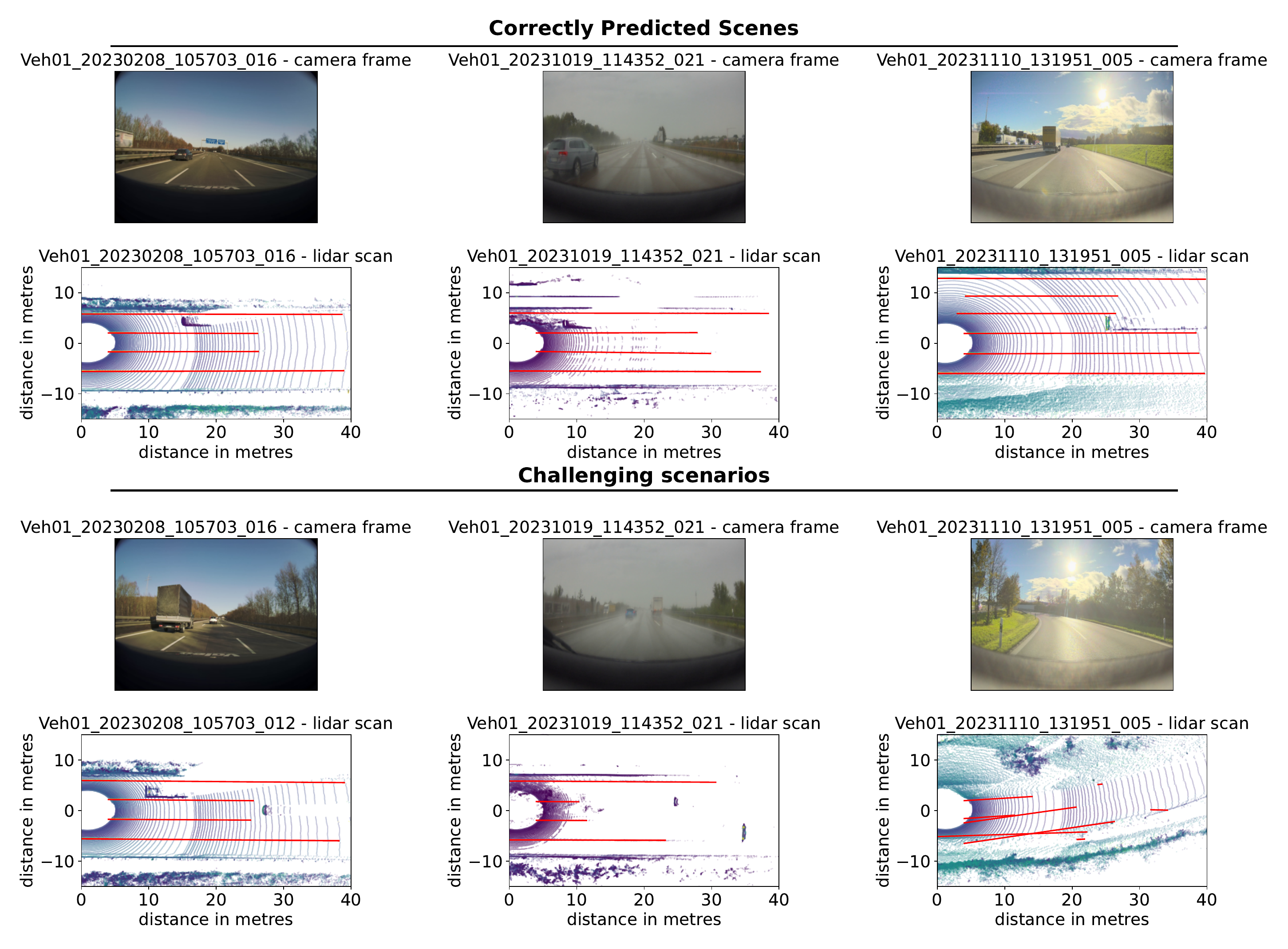}
    \caption{Qualitative results showing successful (top row) and challenging (bottom row) predictions across different driving scenarios.}
    \label{fig:scenes_examples}
\end{figure}

Challenging scenarios include worn or missing markings (e.g., merging zones), sparse returns in rain, and complex road geometries. Fig.~\ref{fig:scenes_examples} visualizes both successful and failure cases. Top-row scenes show accurate polyline reconstruction; bottom-row examples reveal common failure modes, such as lane confusion or missed markings.

These results suggest the pipeline’s potential for practical deployment, while also highlighting the need for improved handling of occlusions, severe weather, and highly curved lane structures.


\section{Conclusion}

We introduced Pandar128, the largest public LiDAR-based dataset for lane line detection, featuring over 34k annotated LiDAR frames with high-resolution scans, synchronized odometry, and comprehensive calibration data. This enables advanced research in geometric lane modeling, sensor fusion, and long-range perception.

To benchmark this dataset, we proposed SimpleLidarLane, a lightweight baseline pipeline combining BEV segmentation with clustering and polyline fitting. Despite its simplicity, it achieves strong performance across diverse driving scenarios, including challenging weather and illumination conditions.

In addition, we introduced IAM-F1, a novel interpolation-aware metric for evaluating polyline alignment in BEV space. Unlike raster-based metrics, IAM-F1 directly compares sparse polylines, offering improved geometric fidelity and computational efficiency.

Together, the dataset, baseline, and evaluation protocol establish a new foundation for reproducible and interpretable research in LiDAR-based lane detection. All resources are publicly available to encourage further development in this critical area of autonomous driving.

\bibliographystyle{splncs04}
\bibliography{bibliography}

\end{document}